\documentclass[10pt,journal,cspaper,compsoc]{IEEEtran}

\ifCLASSOPTIONcompsoc
  \usepackage[nocompress]{cite}
\else
\fi
\ifCLASSINFOpdf
  \usepackage{epsfig}
  \usepackage{graphicx}
  \graphicspath{{media/}}
\else
\fi
\usepackage[cmex10]{amsmath}
\usepackage{algorithmic}

\usepackage{array}
\usepackage{url}

\usepackage{dsfont}
\DeclareMathOperator*{\argmin}{argmin}
\DeclareMathOperator*{\argmax}{argmax}

\newcommand{\x}{\mathbf{x}}
\newcommand{\W}{\mathbf{W}}
\begin{document}

\title{Active Clustering with Model-Based Uncertainty Reduction}
\author{Caiming~Xiong,~\IEEEmembership{}
David~M.~Johnson,~\IEEEmembership{}
and~Jason~J.~Corso~\IEEEmembership{Member,~IEEE}%
\IEEEcompsocitemizethanks{\IEEEcompsocthanksitem Authors are with the Department 
of Computer Science and Engineering, SUNY at Buffalo, Buffalo, NY, 
14260.%
\IEEEcompsocthanksitem Corresponding author email: \texttt{\small jcorso@buffalo.edu}}%
\thanks{}}

\bstctlcite{IEEEexample:BSTcontrol}

\markboth{Manuscript Submitted to IEEE TPAMI on 7 Feb 2014}%
{}

\IEEEcompsoctitleabstractindextext{%
\begin{abstract} 

Semi-supervised clustering seeks to augment traditional clustering methods by incorporating side information provided via human expertise in order to increase the semantic meaningfulness of the resulting clusters. 
However, most current methods are \emph{passive} in the sense that the side information is provided beforehand and
selected randomly. This may require a large number of constraints, some of which could be redundant, unnecessary, or even detrimental to the clustering results.
Thus in order to scale such semi-supervised algorithms to larger problems it is desirable to pursue an \emph{active} clustering method---i.e. an algorithm that maximizes the effectiveness of the available human labor by only requesting human input where it will have the greatest impact. 
Here, we propose a novel online framework for active semi-supervised spectral clustering that selects pairwise constraints as clustering proceeds, based on the principle of uncertainty reduction.  Using a first-order Taylor expansion, we decompose the expected uncertainty reduction problem into a gradient and a step-scale, computed via an application of matrix perturbation theory and cluster-assignment entropy, respectively.  The resulting model is used to estimate the uncertainty reduction potential of each sample in the dataset.  We then present the human user with pairwise queries with respect to only the best candidate sample.
We evaluate our method using three different image datasets (faces, leaves and dogs),  a set of common UCI machine learning datasets and a gene dataset. The results validate our decomposition formulation and show that our method is consistently superior to existing state-of-the-art techniques, as well as being robust to noise and to unknown numbers of clusters.
\end{abstract}

}
 
\maketitle

\IEEEdisplaynotcompsoctitleabstractindextext

%
\IEEEpeerreviewmaketitle

\section{Introduction}
\label{Intro}

Semi-supervised clustering plays a crucial role in machine learning and computer vision for its ability to enforce top-down structure while clustering \cite{ basu2004probabilistic, li2009constrained,lu2008constrained, xing2003distance, hoi2007learning, ChenZ11}. In these methods, the user is allowed to provide external semantic knowledge---generally in the form of constraints on individual pairs of elements in the data---as \emph{side information} to the clustering process.  These efforts have shown that, \emph{when the constraints are selected well}\cite{davidson2006measuring}, incorporating pairwise constraints can significantly improve the clustering results. 

\begin{figure*}
\begin{center}
   \includegraphics[width=0.8\linewidth]{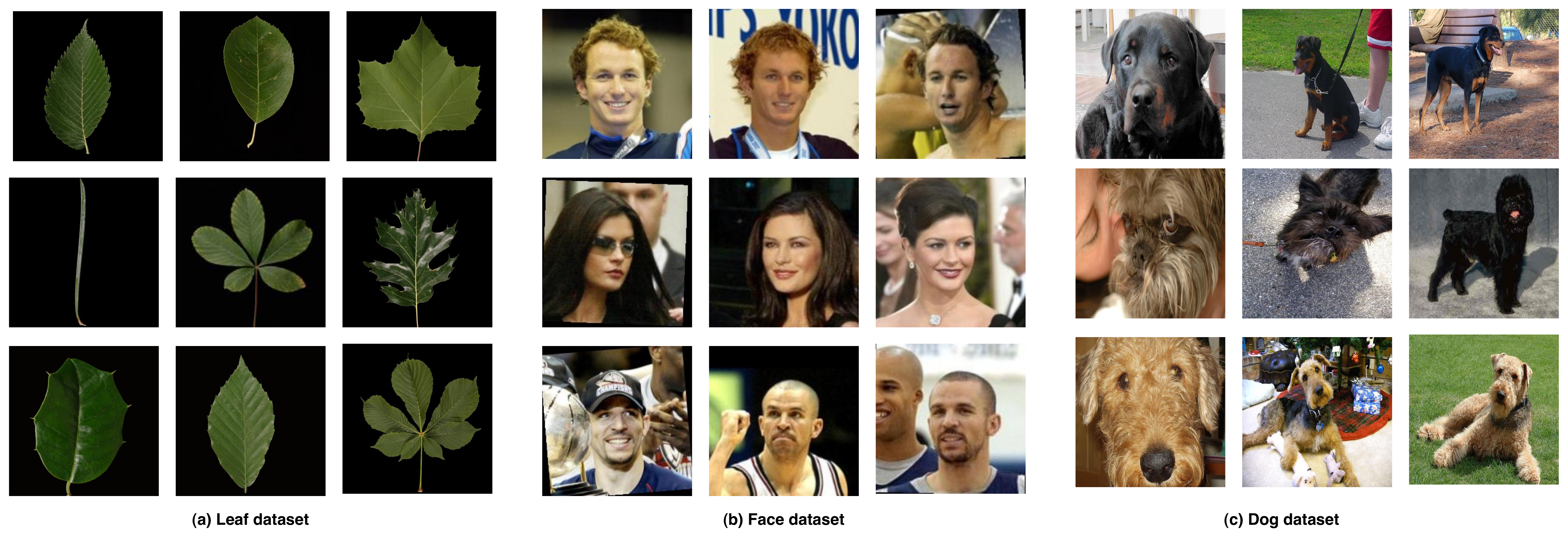}
   \caption{Sample images from three image datasets: (a) Leaves~\cite{leaf}; (b) Faces~\cite{kumar2009attribute}; (c) Dogs~\cite{khosla2011novel}.  \textit{Best viewed in color.}\label{sample1}}
\end{center}
\end{figure*}

In computer vision, there are a variety of domains in which semi-supervised clustering has the potential to be a powerful tool, including, for example,  facial recognition and plant categorization\cite{biswas2013active}.
First, in surveillance videos, there is significant demand for automated grouping of faces and actions: for instance, recognizing that the same person appears at two different times or in two different places, or that someone performs a particular action in a particular location~\cite{zhao2013unsupervised}.  These tasks may be problematic for traditional supervised recognition strategies due to difficulty in obtaining training data---expecting humans to label a large set of strangers' faces or categorize every possible action that might occur in a video is not realistic.  However, a human probably \emph{can} reliably determine whether two face images are of the same person~\cite{biswas2013active} or two recorded actions are similar, making it quite feasible to obtain pairwise constraints in these contexts.  

The problem of plant identification is similar in that even untrained non-expert humans~\cite{kumar2012leafsnap} (for instance, on a low-cost crowd-sourcing tool such as Amazon's Mechanical Turk\cite{buhrmester2011amazon}) can probably generally determine if two plants are the same species, even if only an expert could actually provide a semantic label for each of those images.  Thus, non-expert labor, in conjunction with semi-supervised clustering, can reduce a large set of uncategorized images into a small set of clusters, which can then be quickly labeled by an expert.  The same pattern holds true in a variety of other visual domains, such as identifying animals or specific classes of man-made objects, as well as non-visual tasks such as  document clustering \cite{huang2009active}.


However, even when using relatively inexpensive human labor, any attempt to apply semi-supervised clustering methods to large-scale problems must still consider the cost of obtaining large numbers of pairwise constraints.  As the number of possible constraints is quadratically related to the number of data elements, the number of possible user queries rapidly approaches a point where only a very small proportion of all constraints can feasibly be queried.  Simply querying random constraint pairs from this space will likely generate a large amount of redundant information, and lead to very slow (and expensive) improvement in the clustering results.  Worse, Davidson et al. \cite{davidson2006measuring} demonstrated that poorly chosen constraints can in some circumstances lead to worse performance than no constraints at all. 

To overcome these problems, our community has begun exploring \emph{active} constraint selection methods \cite{basu2004active, yfu2011, mallapragada2009active,xiong2012online, xiongspectral, XiongJC13 ,xu2005active, hoi2008active, wangactive, biswas2011large,wauthier2012active, sxiong}, which allow semi-supervised clustering algorithms to intelligently select constraints based on the structure of the data and/or intermediate clustering results. These active clustering methods can be divided into two categories: sample-based and sample-pair-based. 

The sample-based methods first select samples of interest, then query pairwise 
constraints based on the selected sample \cite{basu2004active,  
mallapragada2009active, xiong2012online}. Basu et al. \cite{basu2004active} propose offline (i.e., not based on intermediate clustering results)
active k-means clustering based on a two-stage process that first explores the problem space and performs user queries to initialize and grow sets of samples with known cluster assignments, and then extracts a large constraint set from the known sample sets and does semi-supervised clustering.
 Mallapragada et al.~\cite{mallapragada2009active} present another active k-means
method based on a min-max criterion, which also utilizes an initial ``exploration'' phase to determine the basic cluster structure.
We have also previously proposed two different sample-based active clustering methods~\cite{XiongJC13, xiong2012online}. This paper represents an improvement and extension of these works.

By contrast, the sample-pair-based methods \cite{xu2005active, 
hoi2008active, wangactive,biswas2011large,wauthier2012active} directly seek pair constraints to query. Hoi et 
al.~\cite{hoi2008active}  
provide a min-max framework to identify the most informative pairs for 
non-parametric kernel learning and provide encouraging results. However, the 
complexity of that method (which requires the solution of an approximate semidefinite programming (SDP) 
problem) is high, limiting both the size of the data and the number of 
constraints that can be processed. Xu et al.\cite{xu2005active} and Wang and Davidson\cite{wangactive} both propose active spectral clustering methods, but both of them are designed for two-class problems, and poorly suited to the multiclass case. Most recently, Biswas and Jacobs \cite{biswas2013active} propose a
method that seeks pair constraints that maximize the \emph{expected change} in the 
clustering result.  This proves to be a meaningful and useful criterion, but the proposed method requires recomputing potential clustering results many times for each sample-pair selected, and is thus slow.

Both types of current approaches suffer from drawbacks:  most current sample-based methods are offline algorithms that select all of their constraints in a single selection phase before clustering, and thus cannot incorporate information from actual clustering results into their decisions.  Most pair-based methods are online, but have very high computational complexity due to the nature of the pair selection problem (i.e. the need to rank $\mathbf{O}(n^2)$ candidate pairs at every iteration), and thus have severely limited scalability.

In this paper, we  overcome the limitations of existing methods and propose a novel sample-based active spectral clustering framework using \emph{certain-sample sets} that performs efficient and effective sample-based constraint selection in an online iterative manner (certain-sample sets are sets containing samples with known pairwise relationships to all other items in the certain-sample sets). In each iteration of the algorithm, we find the sample that will yield the greatest predicted \textbf{reduction} in clustering \textbf{uncertainty}, and generate pairwise queries based on that sample to pass to the human user and update the certain-sample sets for clustering in the next iteration. Usefully, under our framework the number of clusters need not be known at the outset of clustering, but can instead be discovered naturally via human interaction as clustering proceeds (more details in Section~\ref{overview}).

In our framework, we refer to the sample that will yield the greatest expected uncertainty reduction as the \textbf{most informative sample}, and our active clustering algorithm revolves around identifying and querying this sample in each iteration.  In order to estimate the uncertainty reduction for each sample, we propose a novel approximated first-order model which decomposes expected uncertainty reduction into two components: a gradient and a step-scale factor.  To estimate the gradient, we adopt matrix perturbation theory to approximate the first-order derivative of the eigenvectors of the current similarity matrix with respect to the current sample.  For the step-scale factor we use one of two entropy-based models of the current cluster assignment ambiguity of the sample.  We describe our framework and uncertainty reduction formulation fully in Section~\ref{overview}.
 
We compare our method with baseline and state-of-the art active clustering techniques on three image datasets (face images~\cite{kumar2009attribute}, leaf images~\cite{leaf} and dog images~\cite{khosla2011novel}), a set of common UCI machine learning datasets~\cite{Bache+Lichman:2013} and a gene dataset~\cite{cho1998genome}. Sample images from each set can be seen in Figure~\ref{sample1}.
Our results (see Section~\ref{sec:results}) show that given the same number of pairs queried, our method performs significantly better than existing state-of-the-art techniques.

\section{Background and Related Work}
\textbf{What is clustering uncertainty?} Clustering methods are ultimately built on the relationships between pairs of samples.  Thus, for any clustering method, if our data perfectly reflects the ``true'' relationship between each sample-pair, then the method should always achieve the same perfect result.  In practice, however, data (and distance/similarity metrics) are imperfect and noisy---the relationship between some pairs of samples may be clear, but for others it is highly ambiguous.  Moreover, some samples may have predominantly clear relationships to other samples in the data, while others may have predominantly ambiguous relationships.  Since our goal in clustering is to make a decision about the assignment of samples to a cluster, despite the inevitable ambiguity, we can view the overall sample-relationship ambiguity in the data as the uncertainty of our clustering result.

We then posit that the advantage of semi-supervised clustering is that it eliminates some amount of uncertainty, by removing all ambiguity from pair relationships on which we have a constraint.  It thus follows that the goal of active clustering should be to choose constraints that \emph{maximally} reduce the total sample-assignment uncertainty.  In order to achieve this, however, we must somehow measure (or at least estimate) the uncertainty contribution of each sample/sample-pair in order to choose the one that we expect to yield the greatest reduction. In this paper, we propose a novel first-order model with matrix perturbation theory and the concept of local entropy to the contribution of selected sample, more details in Section~\ref{sec:as}.

\textbf{Why \emph{sample}-based uncertainty reduction?} There are two main reasons 
for proposing a sample-based approach rather than a sample-pair-based one.  First, an uncertain 
pair may be uncertain either because it contains one uncertain sample or because it contains 
\emph{two} uncertain samples.  In the latter case, because the constraint between 
these samples will not extrapolate well beyond them, it yields limited information.  Second, due to the presence of $n^2$ pair constraints for every $n$ samples, pair selection has an inherently higher complexity, which limits the scalability of a pair-based approach. 

\begin{figure*}
  \centering
   \includegraphics[width=0.8\linewidth]{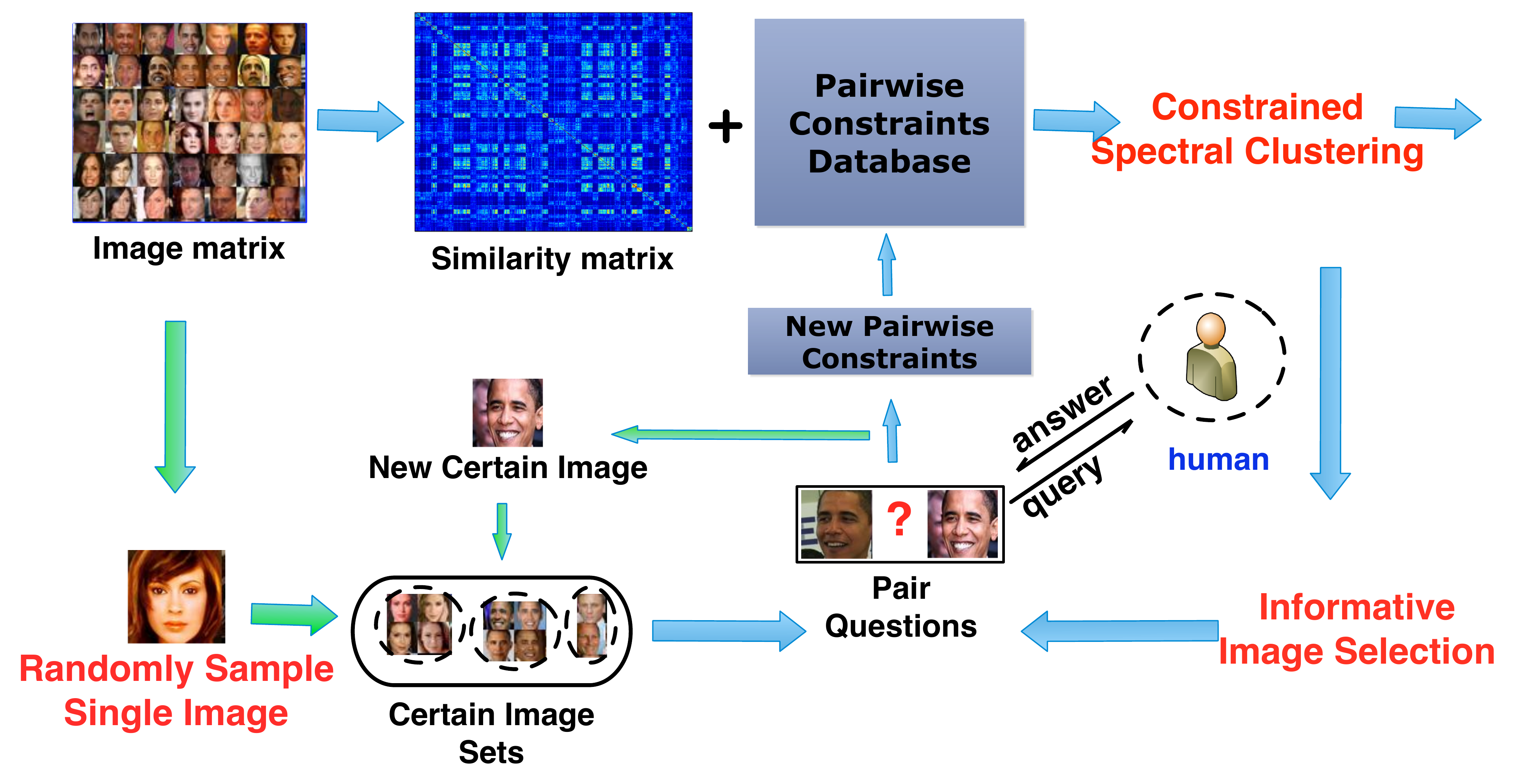}
   \caption{Pipeline of our active clustering framework as applied to image clustering. We iteratively choose a maximally informative image, then select and query new pairwise constraints based on the chosen image, update the certain image sets, and refine the clustering results before returning to select a new most informative image.
   \label{frame} \label{fig:framework}}
\end{figure*} 

\textbf{Relation to active learning.}
Active query selection has previously seen extensive use in the field of active learning \cite{settles2010active, beygelzimer2010agnostic }. Huang et al.~\cite{huang2011active} and Jain and Kapoor~\cite{jain2009active},  for example, both offer methods similar to ours in that they select and query uncertain samples. However, in active learning algorithms the oracle (the human) needs to know the class label of the queried data point.  This approach is not applicable to many semi-supervised clustering problems, where the oracle can only give reliable feedback about the relationship between pairs of samples (such as the many examples we offered in the Section~\ref{Intro}). Though we implicitly label queried samples by comparing them to a set of exemplar samples representing each cluster, we do so strictly via pairwise queries.

Additionally, for the sake of comparison we begin our experiments with an exploration phase that identifies at least one member of each cluster (thus allowing us to treat the clusters we are learning as ``classes'' as far as the active learning algorithms are concerned), but in real data this may not be a reliable option.  There may simply be too many clusters to fully explore them initially, new clusters may appear as additional data is acquired, or certain clusters may be rare and thus not be encountered for some time.  In all of these cases, our active clustering framework can adapt by simply increasing the number of clusters.  In contrast, most active learning methods must be initialized with at least one sample of each class in the data, and do not allow online modification of the class structure.

\section{Active Clustering Framework With Certain-Sample Sets}
\label{overview}
Recall that ``certain-sample sets'' are sets such that any two samples in the same certain-sample set are constrained to reside in the same cluster, and any two samples from different certain-sample sets are guaranteed to be from different clusters. In the ground-truth used in our experiments, each class corresponds to a specific certain-sample set. In our framework, we use the concept of certain-sample sets to translate a sample selection into a set of pairwise constraint queries. 

Given the data set $X=\{x_1,x_2,\cdots,x_n\}$, denote the corresponding pairwise similarity matrix $\W=\{w_{ij}\}$ (i.e. the non-negative symmetric  matrix consisting of all $w_{ij}$, where $w_{ij}$ is the similarity between samples $x_i$ and $x_j$).  Similarity is computed in some appropriate, problem-specific manner.

Here, we also denote the set of certain-sample sets $\mathcal{Z}=\{Z_1,\cdots,Z_m\}$,  where $Z_i$ is a certain-sample set such that $Z_i\subset X$ and $Z_i \cap Z_j = \emptyset$ for all $j$, and define an sample set $\mathcal{O} = \bigcup_i Z_i$ containing \emph{all} current certain sample. Our semantic constraint information  is contained in the set $Q$, which consists of all the available pairwise contraints.  Each of these constraints may be either ``must-link'' (indicating that two samples belong in the same semantic grouping/certain-sample set) or ``cannot-link'' (indicating that they do not).  To initialize the algorithm, we randomly select a single sample $x_i$ such that $Z_1=\{x_i\}$ with $\mathcal{Z}=\{Z_1\}$, $\mathcal{O}= \{x_i\}$ and $Q=\emptyset$.  As $\mathcal{Z}$, $\mathcal{O}$ and $\mathcal{Q}$ change over time, we use the notation $(\cdot)^{t}$ to indicate each of these and other values at the $t^{th}$ iteration.

Assuming we begin with no pairwise constraints,  if the number of clusters in the problem is not known, set the initial cluster number $n_c=2$, otherwise set it to the given number. We then propose the following algorithm (outlined in Figure ~\ref{frame}, more details for each step can be found in Sections \ref{sssc}--\ref{mic}):

\begin{itemize}
\item[1] \textbf{Initialization:} randomly choose a single sample $x_i$, assign $x_i$ to the first certain set $Z_1$ and initialize the pairwise constraint set $\mathcal{Q}$ as the empty set.
\item[2] \textbf{Constrained Spectral Clustering:} cluster all sample into $n_c$ groups using the raw data $X$ plus the current pairwise constraint set $Q$.
\item[3] \textbf{Informative Sample Selection:} choose the most informative sample $x_j$ based on our uncertainty reduction model.
\item[4] \textbf{Pairwise Queries:} present a series of pairwise queries on the chosen sample $x_j$ to the oracle until we have enough information to assign the sample $x_j$ to a certain-sample set $Z_k$ (or create a new certain set for the chosen sample).
\item[5] \textbf{Repeat:} steps 2-4 until the oracle is satisfied with the clustering result or the query budget is reached.
\end{itemize}

It should be noted that, aside from the ability to collect maximally useful constraint information from the human, this algorithm has one other significant advantage: the number of clusters in the problem need not be known at the outset of clustering, but can instead be discovered naturally via human interaction as the algorithm proceeds.  Whenever the queried pairwise constraints result in the creation of a new certain-sample set, we increment $n_c$ to account for it.  This allows the algorithm to naturally overcome a problem faced not just by other active clustering (and active learning) methods, but by clustering methods in general, which typically require a parameter controlling either the size or number of clusters to generate.  This is particularly useful in the image clustering domain, where the true number of output clusters (e.g. the number of unique faces in a dataset) is unlikely to be initially available in any real-world application. We have conducted experiments to evaluate this method of model selection; the results, which are encouraging, are presented in Section~\ref{nonumer}.

Recalling the steps of our framework, from here we proceed iteratively through the three main computational steps: clustering with pairwise constraints, informative sample selection and querying pairwise constraints.  We now describe them.

 \subsection{Spectral clustering with pairwise constraints}
\label{sssc} 

Spectral clustering is a well-known unsupervised clustering method\cite{ng2002spectral}. Given the $n\times n$ symmetric similarity matrix $\W$, denote the Laplacian matrix as $\mathbf{L}=\mathbf{D}-\W$, where $\mathbf{D}$ is the degree matrix such that $\mathbf{D} = \{d_{ij}\}$, where $d_{ij}=\sum_k \mathbf{W}_{ik}$ if $i=j$ and 0 otherwise. Spectral clustering partitions the $n$ samples into $n_c$ groups by performing k-means on the first $n_c$ eigenvectors of $L$. The $n_c$ eigenvectors can be found via:
\begin{alignat}{3}
\mathbf{v} 	& = &&\quad\argmin_{\mathbf{v} } \mathbf{v} ^TL\mathbf{v} \nonumber 						\\
	& = &&\quad \argmin_\mathbf{v}  \sum_{ij} w_{ij}\|\mathbf{v}_i-\mathbf{v}_j\|^2_2 \nonumber		\\
	&     && s.t. \quad \mathbf{v} ^T\mathbf{v}  = I, \mathbf{v} ^T\mathbf{1}=0\enspace.			
\end{alignat}

To incorporate pairwise constraints into spectral clustering, we adopt a simple and effective method called spectral learning \cite{kamvar2003spectral}. Whenever we obtain new pairwise constraints, we directly modify the current similarity matrix $\mathbf{W}^t$, producing a new matrix $\mathbf{W}^{t+1}$. Specifically, the new affinity matrix $W^{t+1}$ is determined via:
 \begin{itemize}
 \item Set $\mathbf{W}^{t+1} = \mathbf{W}^t$.
 \item For each pair of must-linked samples $(i,j)$ assign the values $\mathbf{W}^{t+1}_{ij}=\mathbf{W}^{t+1}_{ji}=1$.
 \item For each pair of cannot-linked samples $(i,j)$ assign the value $\mathbf{W}^{t+1}_{ij}=\mathbf{W}^{t+1}_{ji}=-1$.
 \end{itemize}
We then obtain the new Laplacian matrix $L^{t+1}$ and proceed with the standard spectral clustering procedure.

\subsection{Informative sample selection} 
\label{sec:as}
In this section, we formulate the problem of finding the most informative sample as one of uncertainty reduction.  We ultimately develop and discuss a model for this uncertainty reduction in Section~\ref{sec:as1}.

Define the uncertainty of the dataset in the $t^{th}$ iteration to be conditioned on the current updated similarity matrix $\mathbf{W}^t$ and the current certain-sample set $\mathcal{O}^t$. Thus the uncertainty can be expressed as $\mathbf{U}(X|\mathbf{W}^t,\mathcal{O}^t)$.  Therefore our objective function for sample selection is as follows:
\begin{align}
\x_j^{*}&= \argmax_{\x_j \in X} \Delta \mathbf{U}(\x_j)
            \enspace. \nonumber\\
 \Delta \mathbf{U}(\x_j)&=\mathbf{U}(X|\mathbf{W}^t,\mathcal{O}^t)
- \mathbf{U}(X|\mathbf{W}^t,\mathcal{O}^t \cup \{x_j\}) \label{ob} \enspace.
\end{align}

To the best of our knowledge, there is no direct way of computing uncertainty on the data.  In order to optimize this objective function, we  consider that querying pairs to make a chosen sample ``certain'' can remove ambiguity in the clustering solution and thus reduce the uncertainty of the dataset as a whole. So the expected change in the clustering solution that results from making the chosen sample ``certain'' can be considered as the uncertainty contribution of the sample as a result of selecting and querying that sample.



Thus, we seek samples that will have the greatest impact on the clustering solution.  One strategy for finding these constraints (employed in Biswas and Jacobs~\cite{biswas2013active}, though with sample-pairs rather than samples) is to estimate the likely value of a constraint (i.e. cannot- or must-link) and simulate the effect that constraint will have on the clustering solution.  However, this approach  is both unrealistic (if the answers given by the oracle could be effectively predicted, the oracle would not be needed) and computationally expensive (in the worst case requiring a simulated clustering operation for each possible constraint at each iteration of the active clusterer). 

Thus, we adopt a more indirect method of estimating the impact of a sample query based on matrix perturbation theory and local entropy of selected sample. We present the details of our method in Section~\ref{sec:as1}.

\subsection{Sample-based pairwise constraint queries}
\label{mic}
Before presenting our model for informative sample selection, we briefly describe how we use the selected sample. Because our active selection system is sample-based and our constraints pair-based, once we have selected the most informative sample we must then generate a set of pairwise queries related to that sample.  Our goal with these queries is to obtain enough information to add the sample to the correct certain-sample set.  We generate these queries as follows.

First, for each certain set $Z_j$, choose the single sample within the set that is closest to the selected sample $\x_i$ ($ \x_l=\argmax_{\x_l\in Z_j} w_{il} $) and record this sample.  

Second, since there are $m$ certain sets, we will have recorded $m$ sample and similarity values.  We sort these samples based on their corresponding similarity, then, in order of descending similarity, query the oracle for the relation between the selected sample $\x_i$ and $\x_l$ until we find a must-link connection.  We then add $\x_i$ into the certain-sample set containing that $x_l$. If all of the relations are cannot-link, we create a new certain-sample set $Z_{m+1}$ and add $\x_i$ to it. This new certain set $Z_{m+1}$ is then added to $\mathcal{Z}$.  Regardless, $\mathcal{O}$ is correspondingly updated by adding $\x_i$. If the value of $m$ after querying is greater than $n_c$, we also update $n_c$ to reflect the newly discovered ground-truth cluster.  Since the relation between the new sample and all certain sets in $\mathcal{Z}$ is known, we can now generate new pairwise constraints between the selected sample $\x_i$ and all samples in $\mathcal{O}$ without submitting any further queries to the human.

\section{Uncertainty Reduction Model for Informative Sample Selection}
\label{sec:as1}
As described in Section~\ref{sssc}, we use spectral learning \cite{kamvar2003spectral} as our clustering algorithm. In spectral learning \cite{kamvar2003spectral}, the clustering result arises from the values of the first $n_c$ eigenvectors of current similarity matrix.  Therefore, the impact of a sample query on the clustering result can be approximately measured by estimating its impact on $V^{n_c}$ (the first $n_c$ eigenvectors $v_k$):
\begin{align}
\Delta \mathbf{U}(x_j) \approx& \Delta \mathbf{V}^{n_c}(x_j) \nonumber\\
=&\sum_{k=0}^{n_c} \Delta v_k(x_j) \enspace.
\end{align}

In order to measure $ \Delta \mathbf{V}^{n_c}(x_j)$, based on a first-order Taylor expansion, we decompose the change in the eigenvectors into a gradient and a step-scale factor:
\begin{align}
\label{decomp}
\Delta \mathbf{V}^{n_c}(x_j) =& \frac{\partial V^{n_c}(x_j)}{\partial H(x_j)} \Delta H(x_j) \enspace,
\end{align}
where $H(x_j)$ represents the assignment-ambiguity of $x_j$,  and $\Delta H(x_j)$ represents the reduction in this ambiguity after querying $x_j$. $\frac{\partial V^{n_c}(x_j)}{\partial H(x_j)}$ is a first-order derivative of the changes in the eigenvectors as a result of this ambiguity reduction.  We describe how to estimate this gradient and ambiguity reduction in Sections \ref{slope} and \ref{lum}, respectively.

%
%

\subsection{Estimating the gradient of the uncertainty reduction}
\label{slope}

In order to solve (\ref{decomp}) we must first evaluate $\frac{\partial V^{n_c}(x_j)}{\partial H(x_j)}$.  
Since, we know that in spectral learning (Section~\ref{sssc}) the information obtained from the oracle queries is expressed via changes in the similarity values for the queried point contained in $\mathbf{W}^t$.  Given this, changes in ambiguity are always mediated by changes in $\mathbf{W}^t$, so we can approximate $\frac{\partial V^{n_c}(x_j)}{\partial H(x_j)}$ via
\begin{align}
\frac{\partial V^{n_c}(x_j)}{\partial H(x_j)} \approx \frac{\partial V^{n_c}(x_j)}{\partial \mathbf{W}^t_{x_j}} \enspace,
\end{align}
where $\partial \mathbf{W}^t_{x_j}$ represents an incremental change in the similarity values of sample $x_j$.

Thus, we must begin by computing $\frac{\partial V^{n_c}(x_j)}{\partial \mathbf{W}^t_{x_j}}$ for each $x_j$, for which we propose a method based on matrix perturbation theory \cite{stewart1990matrix}. First note that the graph Laplacian at iteration $t$ can be fully reconstructed from the eigenvectors and corresponding eigenvalues via $L^{t}= \sum_{i=1}^{n} \lambda_i v_iv_i^T$.
Then, given a small constant change in a similarity value $w_{jk}^t$, the first-order change of the eigenvector $v_i$ can be calculated as:
\begin{align}
\frac{dv_i}{dw_{jk}^t} = \sum_{p\not=i}\frac{v_i^T\left(\partial L^{t}/\partial w_{jk}^t\right)v_p}{\lambda_i-\lambda_p} v_p\label{pair}
\end{align}
Note that $\partial L^{t}/\partial w_{jk}^t=(e_j-e_k)(e_j-e_k)^T$, where $e_q$ is the $n$-length indicator vector of index $q$. 

For the chosen sample $x_j$ we take $n_c$ samples $X_{n_c}=\{x_{j_1},x_{j_2},\cdots,x_{j_{n_c}}\}$, one sampled from each certain set $Z_i\in \mathcal{Z}$.  If we decide to query the oracle for $x_j$, the relation of $x_j$ to each sample in $X_{n_c}$ will become known, and the corresponding $w_{jk}^t$ in $W^t$ will be updated during spectral learning.  Therefore, to estimate the influence of sample $x_j$ on the gradient of the eigenvectors, we can simply sum the influences of the relevant $w_{jk}^t$ values based on Eq.~\ref{pair}.  We thus define our approximate model for the derivative of uncertainty reduction as:
\begin{align}
\frac{\partial V^{n_c}(x_j)}{\partial H(x_j)} &\approx \sum_{i=1}^{n_c} \left|\sum_{x_k\in X_{n_c}}\frac{dv_i}{dw_{jk}^t}\right| \nonumber \\
&=\sum_{i=1}^{n_c} \left|\sum_{x_k\in X_{n_c}} \sum_{p\not=i}\frac{v_i^T[\partial L^{t}/\partial w_{jk}^t]v_p}{\lambda_i-\lambda_p}v_p\right| \enspace.
\end{align}

Note that we operate only over a subset of certain samples in order to both save on computation and avoid redundancy.  We could simply use the entirety of $\mathcal{O}$ in place of $X_{n_c}$, but this would likely distort the results.  Intuitively, the effect of a must-link constraint is to shift the eigenspace representations of the two constrained samples together.  The samples in a certain set should thus have very similar eigenspace representations, so we expect additional constraints between them and $x_j$ to have diminishing returns.  

\subsection{Estimating the step scale factor for uncertainty reduction}
\label{lum}

The second component of our uncertainty reduction estimation is $\Delta H(x_j)$---the change in the ambiguity of the sample $x_j$ as a result of querying that sample.  This component serves as the step scale factor for the gradient $\frac{\partial V^{n_c}(x_j)}{\partial H(x_j)}$.  
According to the assumptions in Section~\ref{mic}, after a sample is queried the ambiguity resulting from that sample is reduced to 0.  This leads to the conclusion that
\begin{align}
\Delta H(x_j) = H(x_j) \enspace.
\end{align}

\begin{figure*}
\begin{center}
   \includegraphics[width=0.8\linewidth]{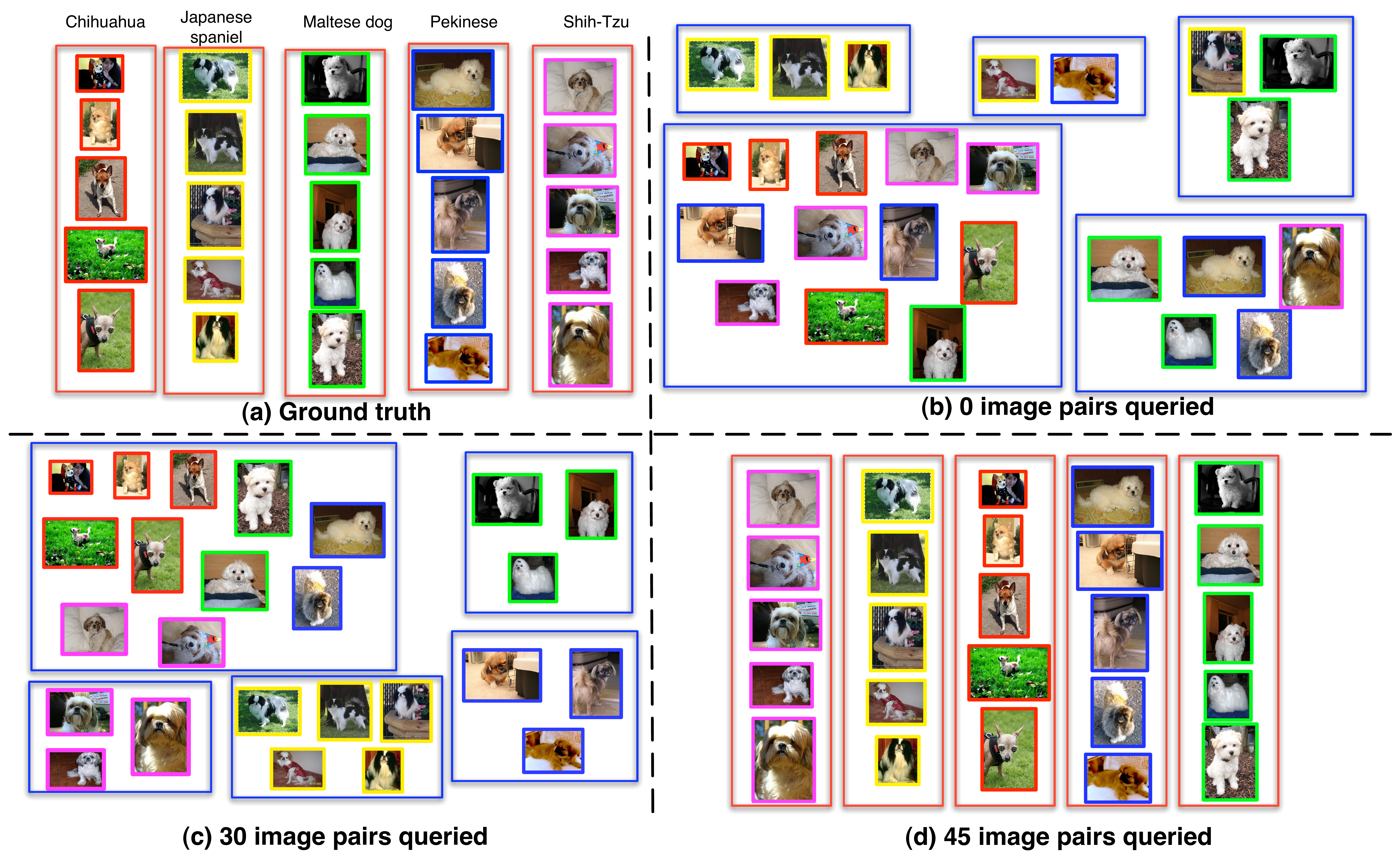}
   \caption{Qualitative results on a small subset of the dog dataset: (a) cluster ground-truth; (b) Initial unconstrained clustering result; (c) result with 30 pairwise constraints queried; (d) result with 45 pairwise constraints queried. \textit{Best viewed in color.}\label{quality}}
\end{center}
\end{figure*}

Therefore, the problem of estimating the change in ambiguity of a sample reduces to the problem of estimating the current ambiguity of that sample.  While this problem still cannot be solved precisely, we present two reasonable heuristics for estimating the ambiguity of a sample.  Both are based on the concept of entropy---specifically, the entropy over probability distributions of local cluster labels (an uncertainty estimation strategy that has shown good results in active learning~\cite{settles2010active}).

\textbf{Nonparametric structure model for cluster probability}
First, consider the current clustering result $C^t=\{c_1,c_2,\cdot,c_{n_c}\}$, where $c_i$ is a cluster and $n_c$ is the number of clusters. We can then define a simple non-parametric model based on similarity matrix $W$ for determining the probability of $x_j$ belonging to cluster $c_i$:
\begin{align}
P( c_i|x_j) = \frac{\sum_{x_l \in c_i}w_{jl}}{\sum_{x_l \in X}w_{jl}}
\end{align}
Because only local nearest neighbors have large similarity values in relation to a given sample, we can use the $k$-nearest neighbors ($k$NN) of each point to efficiently approximate the entropy.  These neighbors need only be computed once,
so this ambiguity estimation process is fast and scalable. In our experiments, we use $k=20$.

\textbf{Parametric model for cluster probability}
Alternately, we can simply use the eigenspace representation of our data produced by the most recent semi-supervised spectral clustering operation to compute a probabilistic clustering solution.  We elect to learn a mixture model (MM) on the embedded eigenspace of the current similarity matrix $W^t$ for this purpose:%
\begin{align}
  p(\x_j|\{\alpha_c\},\theta_c\}) = \sum_{c=1}^{n_c} \alpha_c f(\x_j;\theta_c) \enspace ,
\end{align}
where $\{\alpha_c\}$ are the mixing weights and $(\theta_c\})$ are the component parameters.
Then, the probability of each data point given each cluster $c$ is computed via: 
\begin{align}
P(c|\x_j) = \frac{\alpha_c f(\x_j;\theta_c)}{\sum_i^k \alpha_c f(\x_i;\theta_c)} \enspace.
\end{align}
In our experiments, we assume a Gaussian distribution for each component, yielding a Gaussian Mixture Model (GMM).

\textbf{Entropy-based ambiguity model}
Whether using the parametric or nonparametric cluster probability model, the ambiguity of sample $x_j$ can be defined, based on entropy, as:
\begin{align}
H(x_j) = -\sum_{i=1}^{n_c} P(c_i|x_j) \log P( c_i|x_j)
\end{align}
We then use this value to approximately represent $\Delta H(x_j)$.  In combination with the approximate uncertainty gradient $\frac{\partial V^{n_c}(x_j)}{\partial H(x_j)}$ computed as in Section~\ref{slope}, this allows us to evaluate (\ref{decomp}) and effectively estimate the uncertainty reduction for every point $x_j$.  From there, solving our sample selection objective (\ref{ob}) is a simple $\argmin$ operation. In Figure \ref{quality}, we show a qualitative example on a small subset of dog dataset \cite{khosla2011novel}.  In the top-left, five ground-truth clusters are shown with their dog-breed labels.  The other three panes show clustering with increasing numbers of constraints selected via our method.  Notice how clustering initially (with 0 pair constraints) emphasizes dog image appearance and violates many breed boundaries, whereas with 30 and 45 constraints the clusters are increasingly correct. 

\section{Complexity Analysis}
At each iteration, we must select a query sample from among $\mathbf{O}(n)$ possibilities, applying our uncertainty reduction estimation model to each potential sample. Computing the gradient component of the uncertainty model takes $\mathbf{O}(m{n_c}^2n)$ time for each sample, where $m$ is number of certain sets and $n_c$ is the number of clusters/eigenvectors. $m\leq n_c$, so the complexity of the uncertainty gradient evaluation at each iteration is $\mathbf{O}(n_c^3n^2)$.  Computing all the step scale factors costs $\mathbf{O}(n_ckn)$ (where $k$ is the number of nearest neighbors) if the nonparametric method is used, or $\mathbf{O}(n_c^3n)$ for the parametric method. $k \ll n$, so regardless the total complexity of the active selection process at each iteration is $\mathbf{O}(n_c^3n^2)$.

In order to reduce this cost, we adopt a slight approximation.  In general, the samples with the largest uncertainty reduction will have both a large step scale and a large gradient.  With this mind, we first compute the step scale for each sample (this is cheaper than computing the gradient, particularly if the nonparametric model is used), then only compute the gradient for the $b$ samples with the largest step scales.  Assuming $b \ll n$, this yields an overall complexity of $\mathbf{O}(n_c^3n)$.  Note that all results for our method shown in this paper were obtained using this fast approximation.  Also note that for large data, the cost of the method will generally be dominated by the spectral clustering itself, which is $\mathbf{O}(n^3)$ in the worst case (though potentially significantly cheaper, possibly even $\mathbf{O}(n)$~\cite{chen2011large,yan2009fast}, depending on the eigendecomposition method used and the sparseness of the similarity matrix).
 
\section{Experimental Setup}											
\label{sec:exp}
\subsection{Data} 
We evaluate our proposed active framework and selection measures on three image datasets (leaves, dogs and faces---see Figure \ref{sample1}), one gene dataset~\cite{cho1998genome} and five UCI machine learning datasets~\cite{Bache+Lichman:2013}. 
We seek to demonstrate that our method  is generally workable for different types of data/applications with a wide range of cluster numbers.

\textbf{Face dataset:} all face images are extracted from a face dataset called PubFig~\cite{kumar2009attribute},  which is a large, real-world face dataset consisting of 58,797 images of 200 people collected from the Internet. Unlike most other existing face datasets, these images are taken in completely uncontrolled settings with non-cooperative subjects. Thus, there is large variation in pose, lighting, expression, scene, camera, and imaging conditions. We use two subsets: \textbf{Face-1} (500 images from 50 different peoples) and \textbf{Face-2} (200 images from 20 different people).

\textbf{Leaf dataset:} all leaf images are iPhone photographs of leaves against a monochrome background, acquired through the Leafsnap app~\cite{leaf}. We use the same subset (1042 images from 62 species) as in \cite{biswas2013active}. The feature representations and resulting similarity matrices for the leaf and face datasets are all from \cite{biswas2013active}. 

\textbf{Dog dataset:} all dog images are from the Stanford Dogs dataset \cite{khosla2011novel}, which contains 20,580 images of 120 breeds of dogs. We extract a subset containing 400 images from 20 different breeds (dog-400) and compute the features used in \cite{deng2009imagenet}.  Affinity is measured via a $\chi^2$ kernel.

\begin{table}
\caption{UCI machine learning and gene data sets}
\begin{center}
\begin{tabular}{|l|c|c|c||l|c|c|c|}
\hline
Dataset & Size & Dim. & No. Classes\\
\hline\hline
Balance & 625 & 4 & 3\\
\hline
BUPA Liver Disorders & 345 & 6 & 2\\
\hline
Diabetes & 768 & 8 & 2\\
\hline
Sonar & 208 & 60 & 2\\
\hline
Wine & 178 & 13 & 3\\
\hline
Cho's gene & 307 & 100 & 5\\
\hline
\end{tabular}
\label{data sets}
\end{center}
\end{table}

\textbf{Gene and UCI machine learning datasets:} we choose five datasets from the UCI repository and \textbf{Cho}'s \cite{cho1998genome} gene dataset (details in Table \ref{data sets}). Affinity is measured via a Gaussian kernel.


\subsection{Evaluation protocols} 
We evaluate all cluster solutions via two commonly used cluster evaluation metrics: the Jaccard Coefficient \cite{pang2006introduction} and V-measure \cite{rosenberg2007vmeasure}.

The \textbf{Jaccard Coefficient} is defined by $\text{JCC} = \frac{SS}{SD+DS+SS}$, where:
\begin{itemize}
\item \textbf{SS}: represents the total number of pairs that are assigned to the same cluster in both the clustering results and the ground-truth.
\item \textbf{SD}: represents the total number of pairs that are assigned to the same cluster in the clustering results, but to different clusters in the ground-truth.
\item \textbf{DS}: represents the total number of pairs that are assigned to different clusters in the clustering results, but to the same cluster in the ground-truth.
\end{itemize}


\textbf{V-Measure} is  an alternate
 metric for determining cluster 
correspondence between a set of ground-truth classes $C$ and clusters $K$, which defines entropy-based measures for the completeness 
and homogeneity of the clustering results, and computes the harmonic 
mean of the two. The homogeneity $h$ is:
\begin{align}
 h= \left\{ \begin{array}{ll}
         1 & \mbox{if $H(C,K)=0$};\\
        1-\frac{H(C|K)}{H(C)} & \mbox{else}.\end{array} \right.
\end{align}
where
\begin{align}
H(C|K)=-\sum_{p=1}^{|K|}\sum_{q=1}^{|C|}\frac{a_{qp}}{N}\log \frac{a_{qp}}{\sum_{c=1}^{|C|}a_{qp}}
\end{align}

\begin{align}
H(C)=-\sum_{q=1}^{|C|}\frac{\sum_{p=1}^{|K|}a_{qp}}{n}\log \frac{\sum_{p=1}^{|K|}a_{qp}}{n}
\end{align}

The completeness $c$ is:
\begin{align}
 c= \left\{ \begin{array}{ll}
         1 & \mbox{if $H(K,C)=0$};\\
        1-\frac{H(K|C)}{H(K)} & \mbox{else}.\end{array} \right.
\end{align}
where
\begin{align}
H(K|C)=-\sum_{q=1}^{|C|}\sum_{p=1}^{|K|}\frac{a_{qp}}{N}\log \frac{a_{qp}}{\sum_{p=1}^{|K|}a_{qp}}
\end{align}

\begin{align}
H(K)=-\sum_{p=1}^{|K|}\frac{\sum_{q=1}^{|C|}a_{qp}}{n}\log \frac{\sum_{c=1}^{|C|}a_{qp}}{n}
\end{align}
$a_{qp}$ is the number of data samples that are members of class $q$ and elements of cluster $p$.
The final V-measure for a clustering result is then equal to the harmonic mean of homogeneity and completeness:
\begin{align}
V_{\beta}=\frac{(1+\beta)*h*c}{(\beta*h)+c}
\end{align}
In our case, we weight both measures equally, setting $\beta=1$ to 
yield a single accuracy measure. 

\subsection{Baseline and state-of-the-art methods}
To evaluate our active clustering framework and proposed active constraint selection strategies, we test the following set of methods, including a number of variations on our own proposed method, as well as a baseline and multiple state-of-the-art active clustering and learning techniques.  From this point forward we refer to our proposed method as Uncertainty Reducing Active Spectral Clustering (URASC). The variants of URASC:

\begin{itemize}
\item \textbf{URASC+N:} Proposed model for uncertainty reducing active clustering with gradient and nonparametric step scale estimation.
\item  \textbf{URASC+P:} Proposed model for uncertainty reducing active clustering with gradient and parametric step scale estimation.
\item \textbf{URASC-GO:} Our model without step scale estimation---only the gradient estimation for each sample is used.
\item \textbf{URASC-NO:} Our model without gradient estimation---only the nonparametric step scale is used.
\item \textbf{URASC-PO:} Our model without gradient estimation---only the parametric step scale is used.
\end{itemize}

\begin{figure*}
\begin{center}
   \includegraphics[width=0.82\linewidth]{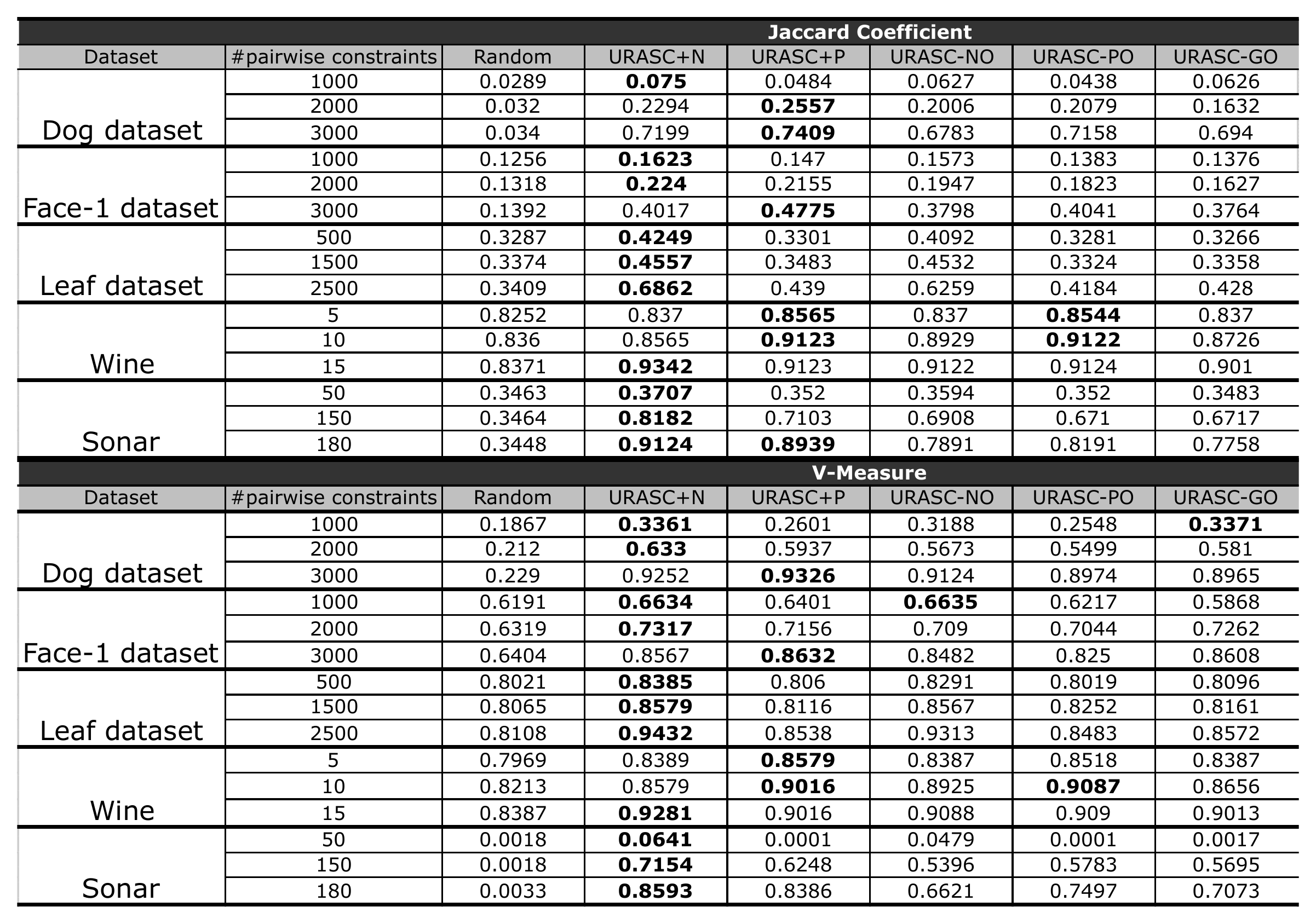}
   \caption{Comparison of variants of our methods against the random baseline.\label{variants}}
\end{center}
\end{figure*}

\begin{figure*}[t]
\begin{center}
   \includegraphics[width=1\linewidth]{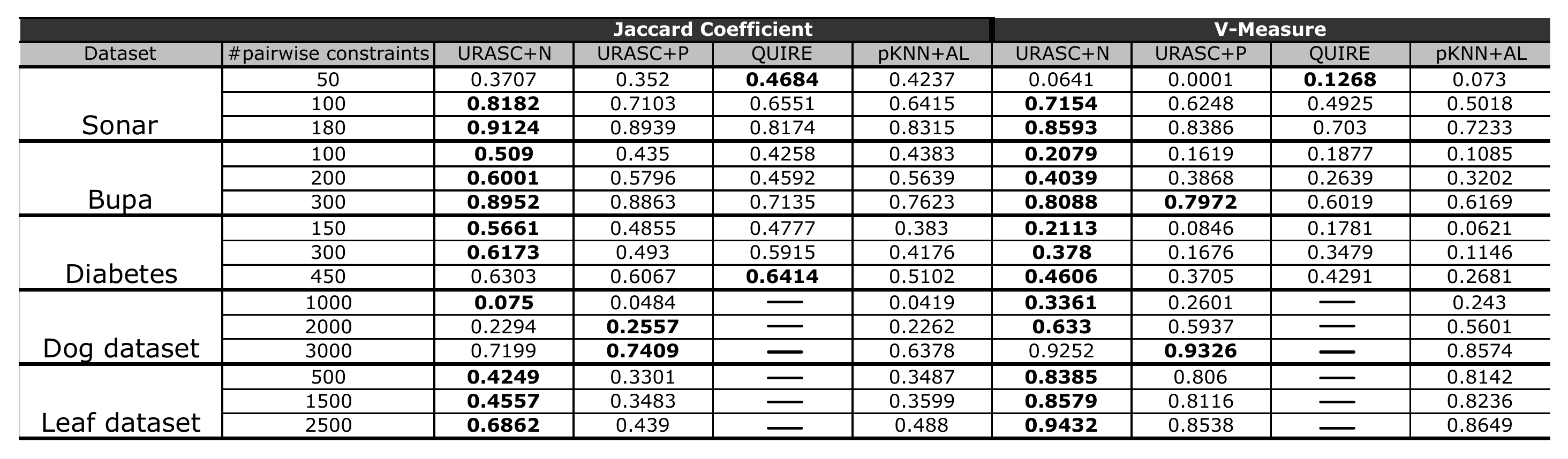}
   \caption{Comparison of our methods against sample-based active learning methods.  \label{activelearning}}
\end{center}
\end{figure*}

\begin{figure*}
\begin{center}
   \includegraphics[width=0.9\linewidth]{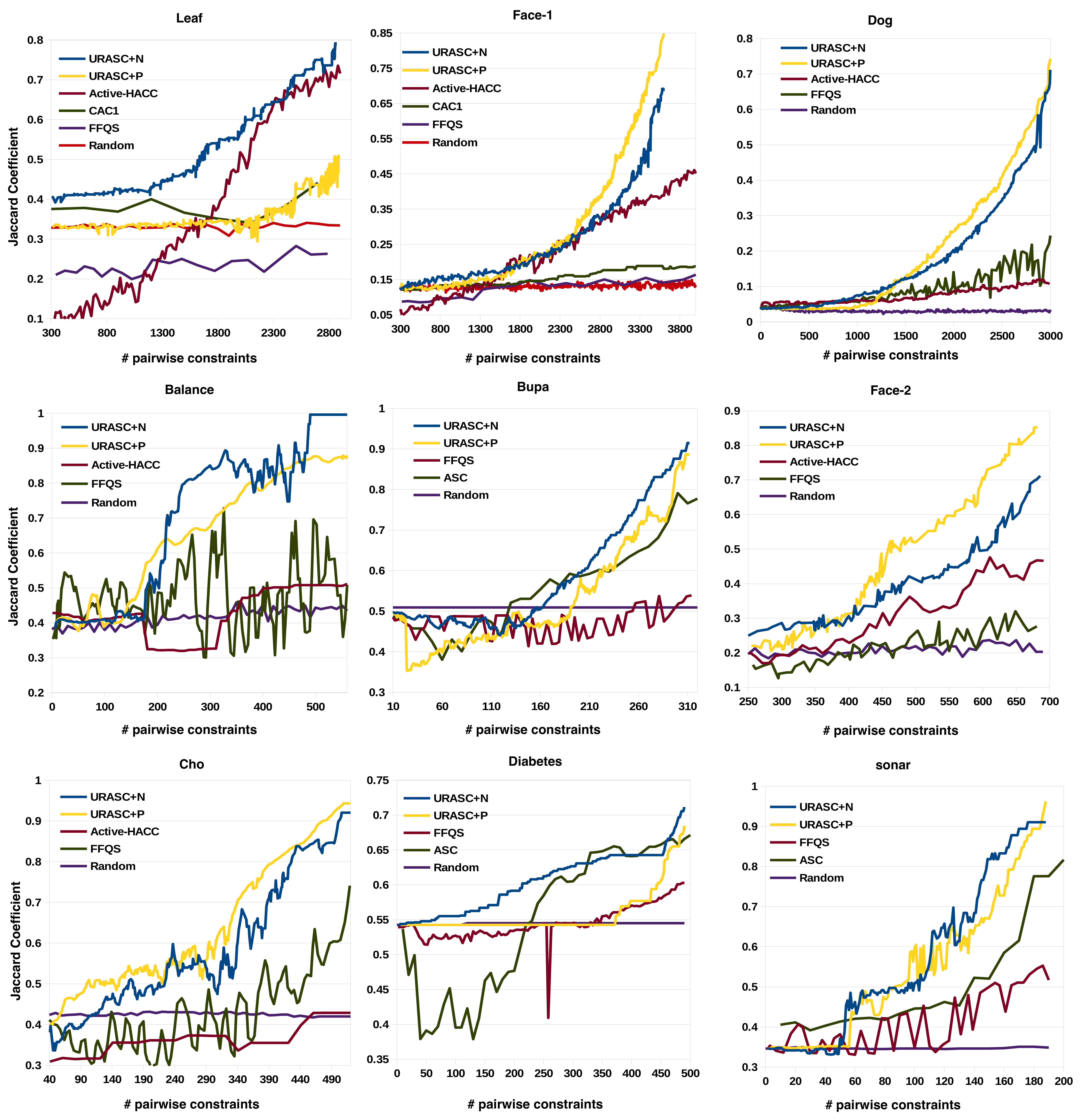}
   \caption{Comparison to state-of-the-art active clustering methods.  y-axis is Jaccard Coefficient score. \textit{Best viewed in color.}\label{activeclustering}}
\end{center}
\end{figure*}

Our baselines and comparison methods include state-of-the-art pair-based active clustering methods and two active learning methods:
\begin{itemize}
\item \textbf{Random}: A baseline in which pair constraints are randomly sampled from the available pool and fed to the spectral learning algorithm.
\item \textbf{Active-HACC:} \cite{biswas2013active} An active hierarchical clustering method that seeks pairs that maximize the expected change in the clustering.
\item \textbf{CAC1:} \cite{biswas2011large} An active hierarchical clustering method that heuristically seeks constraints between large nearby clusters.
\item \textbf{FFQS}  \cite{basu2004active}: An offline active $k$-means clustering method that uses certain-sample sets to guide constraint selection (as in our method), but selects samples to query either through a farthest-first strategy or at random.
\item \textbf{ASC} \cite{wangactive}: A binary-only pair-based active spectral clustering method that queries pairs that will yield the maximum reduction in expected pair value error.  
\item \textbf{QUIRE} \cite{huang2011active}: A binary-only active learning method that computes sample uncertainty based on the informativeness and representativeness of each sample.  We use our certain-sample set framework to generate the requested sample labels from pairwise queries.
\item \textbf{pKNN+AL} \cite{jain2009active}: A minmax-based multi-class active learning method.  Again, we use our framework to translate sample label requests into pairwise constraint queries.
\end{itemize}

\section{Results}
\label{sec:results}
We run our method and its variants on all of the listed datasets and compare against baselines and competing state-of-the-art techniques.  

\subsection{Variant methods and baseline}

In Figure \ref{variants}, we compare our parametric and nonparametric methods, as well as the three ``partial'' URASC procedures, on three image sets and two UCI sets at varying numbers of constraints.  We show results in terms of both Jaccard coefficient and V-measure, and witness similar patterns for each.  In all cases, our parametric and nonparametric methods perform relatively similarly, with the nonparametric having a modest lead at most, but not all, constraint counts.  More importantly, our methods consistently (and in many cases dramatically) outperform the random baseline, particularly as the number of constraints increases.  Our methods always show notable improvement as more constraints are provided---in contrast to the random baseline, which, \emph{at best}, yields minor improvement. Even on the relatively simple wine dataset, it is clear that randomly selected constraints yield little new information.

Finally, we note that our ``complete'' methods consistently meet or exceed the performance of the corresponding partial methods.  Neither the step-scale-only methods nor the gradient-only method consistently yield better results, but in every case the combined method performs at least on-par with the better of the two, and in some cases significantly better than either (see the sonar results in particular).  These results validate the theoretical conception of our method, showing that the combination of gradient and step-scale is indeed the correct way to represent the active selection problem, and that our method's performance is being driven by the combined information of both terms.

\subsection{Comparison to state-of-the-art active learning methods}

We next compare our methods to two active learning methods, as representatives of other pair-based techniques (Figure \ref{activelearning}).  Here we test on three binary UCI datasets in order to provide a reasonable evaluation of the QUIRE method, which is binary-only.  

At least one (and usually both) of our methods outperforms both QUIRE and pKNN+AL in most cases, only definitively losing out at the very low constraint level on the sonar dataset.  As with the random baseline before, the gap between our methods and the competition generally increases with the number of constraints.  These results suggests that simply plugging active learning methods into a clustering setting is suboptimal---we can achieve better results by formulating a clustering-specific uncertainty reduction objective.

Also notable is the fact that, between the two active learning methods, QUIRE is clearly the superior (at least on problems where it is applicable).   This is significant because, like our method, QUIRE seeks to measure the global impact of a given constraint, while pKNN+AL only models local uncertainty reduction.  This lends further support to the idea that the effect of a given query should be considered within the context of the entire clustering problem, not just in terms of local statistics.

\subsection{Comparison to state-of-the-art active clustering methods}

\begin{figure*}
\begin{center}
   \includegraphics[width=0.85\linewidth]{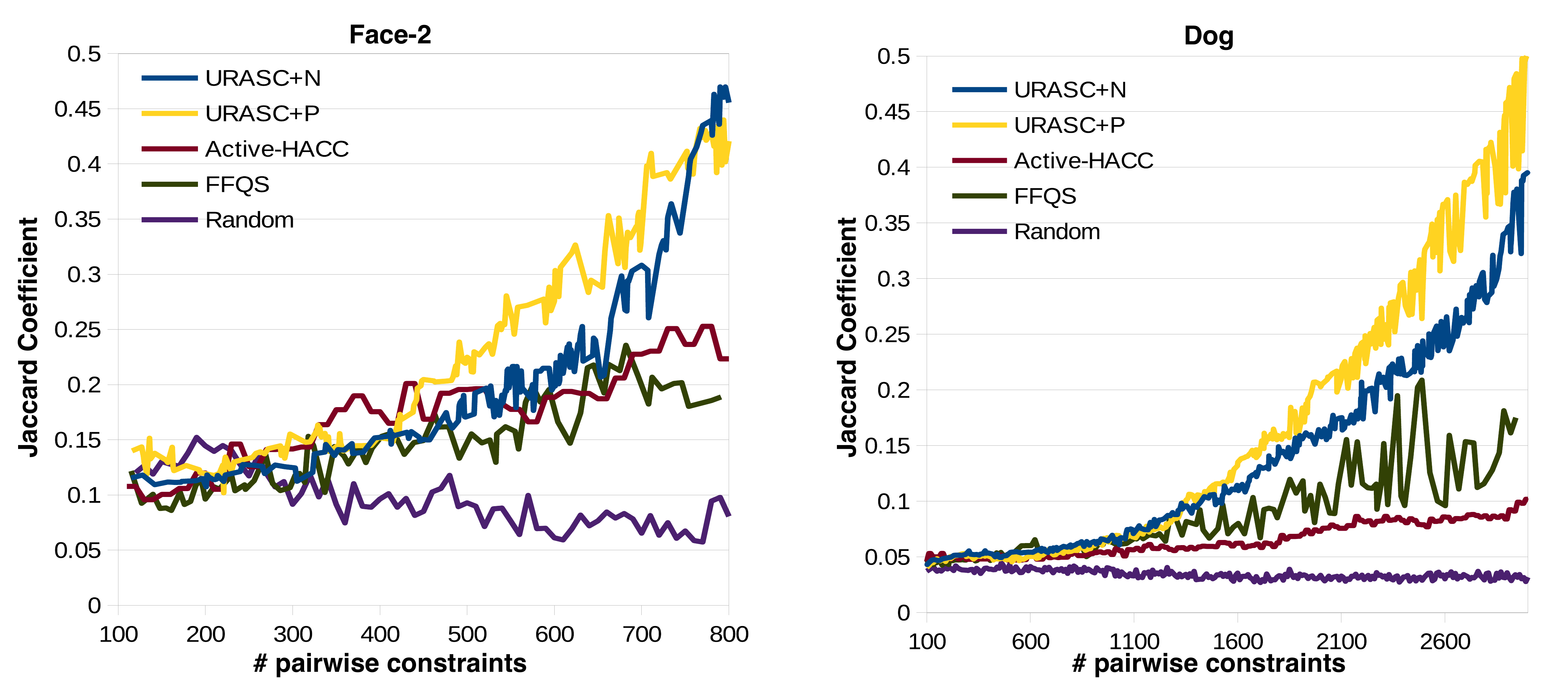}
   \caption{Comparison of our methods against other active clustering methods on two image datasets with a 2\% simulated error rate on the oracle queries. \textit{Best viewed in color.}\label{noise}}
\end{center}
\end{figure*}

Finally, we test our methods against existing active clustering techniques (as well as the random baseline) and represent the results visually in Figure \ref{activeclustering}.  Not all methods appear in all charts because 
ASC\cite{wangactive} is applicable only to binary data.  Once again, our methods present a clear overall advantage over competing algorithms, and in many cases both our parametric and nonparametric methods far exceed the performance of any others (most dramatically on the Dog dataset).  

The only method that comes near to matching our general performance is Active-HACC, which also seeks to estimate the expected change in the clustering as a result of each potential query. However, this method is much more expensive than ours (due to running a large number of simulated clustering operations for every constraint selection) and fails on the Dog dataset.  ASC is also somewhat competitive with our methods, but its binary nature greatly limits its usefulness for solving real-world semi-supervised clustering problems.

Between our two methods, there still appears to be no clear winner, though the nonparametric approach appears to be more reliable given the relative failure of the parametric approach on the Leaf and Diabetes sets.


\subsection{Comparison with noisy input.}  
Our previous experiments are all based on the assumption that the oracle reliably returns a correct ground-truth response every time it is queried.  Previous works in active clustering have also relied on this assumption \cite{basu2004active,yfu2011,mallapragada2009active, xu2005active, wangactive,biswas2011large, wauthier2012active}.  Obviously, this is not, as a general rule, realistic---human oracles may make errors, and in some problems the ground-truth itself may be ambiguous and subjective.  Specifically, for the face and leaf datasets used here, Amazon Mechanical Turk experiments \cite{kumar2009attribute,biswas2013active} have shown that human error is about 1.2\% on face queries and 1.9\% on leaf queries.

Thus, in order to evaluate our active clustering method in a more realistic setting, we performed a set of experiments with a simulated 2\% query error rate on the Face-2 and Dog datasets. We plot the results in Figure~\ref{noise}. We find that, while improvement is noticeably slower and noisier than in the previous experiments, our algorithms still demonstrate a significant overall advantage over other active or passive clustering techniques.  These results also further emphasize the importance of active query selection in general, as with noise added the net effect of the random queries is actually negative.

\subsection{Comparison with unknown numbers of clusters} 
\label{nonumer}

\begin{figure*}
\begin{center}
   \includegraphics[width=0.9\linewidth]{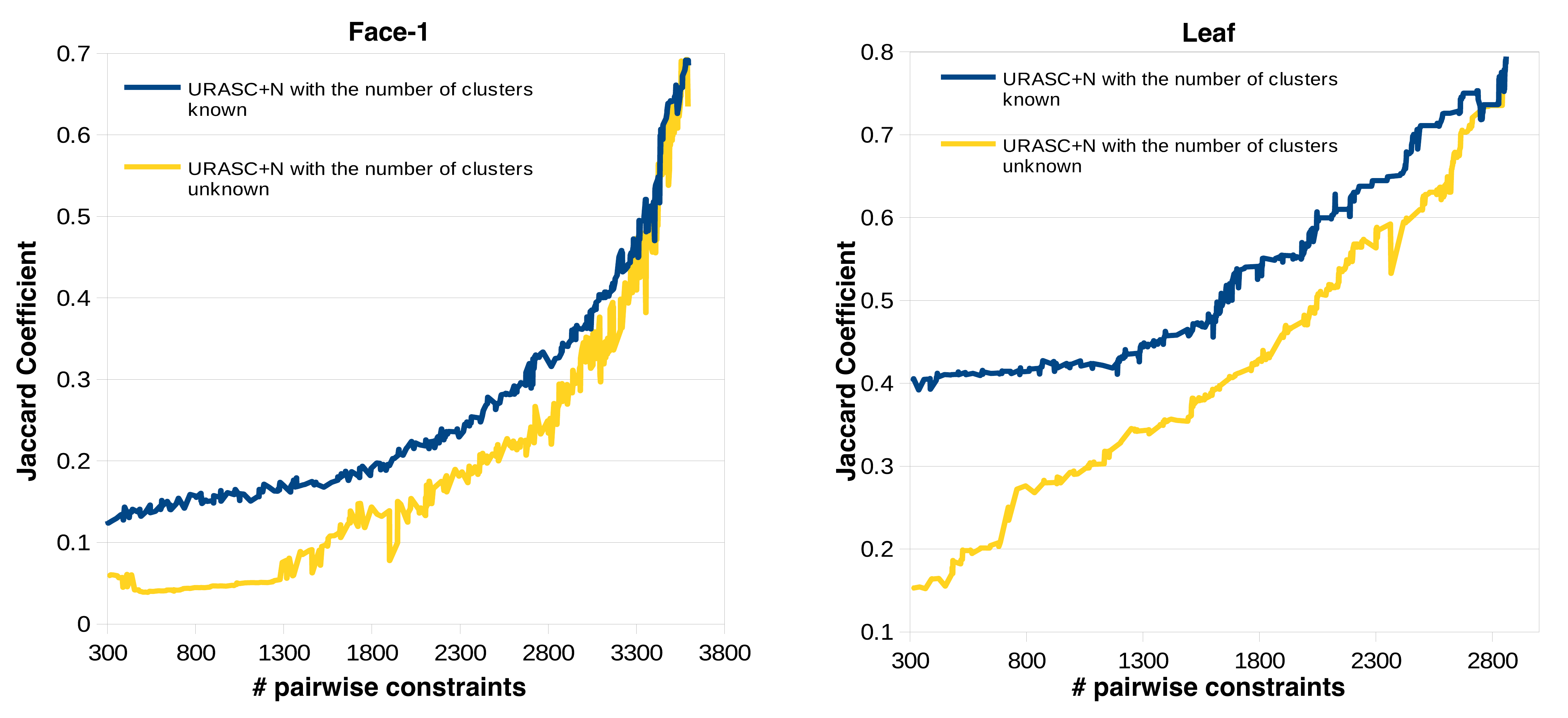}
   \caption{Comparison of URASC+N clustering results with known and (initially) unknown numbers of clusters.  \textit{Best viewed in color.}\label{u_k}}
\end{center}
\end{figure*}

Since one advantage of our method is its ability to dynamically discover the number of clusters based on query results, we analyze how this approach effects performance over time.  We thus run our method on the Face-1 (50 ground-truth clusters) and Leaf (62 ground-truth clusters) datasets, with the number of clusters $k$ initially set to 2, and increasing as new certain-sample sets are discovered. Our results are shown in Figure~\ref{u_k}.   The results are promising, with the unknown-$k$ results initially much lower (as expected), but converging over time towards the known-$k$ results as the cluster structure is discovered.  On both datasets tested, the results appear to eventually become indistinguishable.

\section{Conclusion}
\label{sec:conclusion}
In this paper, we present a novel sample-based online active spectral clustering framework that actively selects pairwise constraint queries with the goal of minimizing the uncertainty of the clustering problem. In order to estimate uncertainty reduction, according to first-order Taylor expansion, we decompose it into a gradient (estimated via matrix perturbation theory) and step-scale (based on one of two models of local label entropy).  We then use pairwise queries to disambiguate the sample with the largest estimated uncertainty reduction. Our experimental results validate this decomposed model of uncertainty and support our theoretical conception of the problem, as well as demonstrating performance significantly superior to existing state-of-the-art algorithms.  Moreover, our experiments show that our method is robust to noise in the query responses and functions well even if the number of clusters in the problem is initially unknown.

One avenue of future research involves reducing the computational burden of the active selection process by adjusting the algorithm to select multiple query samples at each iteration.  The naive approach to this problem---selecting the $k$ most uncertain points---may yield highly redundant information, so a more nuanced technique is necessary.  With this adjustment, this active spectral clustering method could become a powerful tool for use in large-scale online problems, particularly in the increasingly popular crowdsourcing domain.

\section*{Acknowledgements}

We are grateful for the support in part provided through the following grants: NSF CAREER IIS-0845282, ARO YIP W911NF-11-1-0090, DARPA Minds Eye W911NF-10-2-0062, DARPA CSSG D11AP00245, and NPS N00244-11-1-0022.  Findings are those of the authors and do not reflect the views of the funding agencies.

{
\bibliographystyle{IEEEtran}
\bibliography{ace_refs}
}

\end{document}